\documentclass[11pt]{article}
\usepackage[T1]{fontenc}
\usepackage{lmodern}
\usepackage[letterpaper,margin=0.86in]{geometry}
\usepackage{amsmath,booktabs,graphicx,microtype}
\usepackage[hidelinks]{hyperref}
\usepackage[section]{placeins}
\usepackage{url}
\title{Preserving Short Visual Events for Video Question Answering Under Byte and Token Budgets}
\author{Wenji Fu\\[2pt]
\small Southwestern University of Finance and Economics\\[-1pt]
\small \href{mailto:42439069@smail.swufe.edu.cn}{42439069@smail.swufe.edu.cn}\\[-1pt]
\small \href{https://orcid.org/0009-0003-7757-140X}{ORCID 0009-0003-7757-140X}}
\date{November 28, 2026}
\hypersetup{pdftitle={Preserving Short Visual Events for Video Question Answering Under Byte and Token Budgets},pdfauthor={Wenji Fu}}

\begin{document}
\maketitle
\begin{abstract}
Sparse frame selection can miss transient cues needed for video question answering. We evaluate video representations that scan every source frame, store timestamped RGB states, and feed a frozen Qwen3-VL-4B through its native visual interface. We measure transmitted bytes and visual tokens independently. On 280 HERBench questions from ten videos, a 392-token event representation answers 98 correctly, compared with 92 for uniform selection and 79 for full-scan adaptive selection. Flattening local event appearance on the same packet reduces performance to 69 correct, a paired difference of 10.4 percentage points (video-bootstrap 95\% interval, 6.8 to 14.3). On a separate 45-video educational and scientific cohort, regional refresh answers 66 of 222 questions at 392 tokens. Uniform and adaptive 100-frame inputs answer 67 and 66 at 4,900 tokens. On 75 timestamped questions, regional-cache retrieval answers 19 correctly, compared with 15 and 12 from the two 100-frame caches. In a 12-video diagnosis, a dense H.264 cache preserves more measured one-second changes than either 100-frame cache, while answering fewer questions. These results identify settings in which timestamped local appearance is useful and quantify a gap between retaining visual changes and using them to answer questions.
\end{abstract}

\section{Introduction}

Long-video question answering can require evidence from several brief, widely separated events. HERBench explicitly links questions to multiple temporal evidence intervals~\cite{herbench}. Educational tasks also depend on incremental changes in figures, demonstrations, and apparatus~\cite{videomme,videomme2,scivideo}. A small number of whole frames can preserve the scene while missing an intervening object state, annotation, or change in direction.

We study a split pipeline in which a sender observes the source video and a receiver has a frozen vision-language model. The sender can spend bytes on a cache before receiving any questions. The receiver can then spend a separate visual-token budget on answering questions. These two costs can diverge because a cache can hold more states than the model reads for one question. Increasing the number of cached states also does not ensure that a given state survives compression or that the model reads it correctly.

We evaluate three related mechanisms with distinct operating points. An event representation reserves visual tokens for local appearance while retaining a compressed global view. A regional refresh cache retains a time-indexed reconstruction for later lookup. A dense video-codec variant measures how many short changes can be retained at a given physical cache size. Together, the experiments measure question answering, sensitivity to same-packet interventions, and reconstruction fidelity. We include less favorable cohorts and explicit comparisons against frame selection to establish where each effect was observed.

\section{Related work}

Frame selection reduces visual input by keeping representative frames. MaxInfo uses maximum-volume selection in a visual embedding space~\cite{maxinfo}. FrameSelect distributes a frame budget over visual content~\cite{frameselect}. Post-encoder methods such as VidCom$^2$ reduce the number of visual embeddings sent to the language model~\cite{vidcom}. These methods motivate comparisons at the same receiver budget and, where feasible, the same transmitted file size.

Predictive visual coding has direct precedents. AdaCodec learns visual predictive tokens, CoPE-VideoLM aligns codec motion and residuals with image embeddings, and Visual Token Coding allocates visual tokens using intra- and inter-frame structure~\cite{adacodec,cope,vtc}. Our experiments use a frozen RGB receiver and count actual bytes before visual encoding. The distinctive comparison is an intervention after decoding that preserves the physical packet and changes selected event content. We also compare timestamped replay and answer accuracy on the same cached videos.

\section{Methods}

\subsection{Budgets and interface}

Let $X_{0:T-1}$ denote decoded source RGB frames, $P=E(X)$ the sender's serialized packet, and $q$ a question. The receiver answers with a frozen model $f$ from a readout $R$ of $P$, so that $\hat y=f(q,R(P;q))$. The readout uses $q$ only in the timestamped protocol, where it takes explicit times from the question text. Whole-video readout is fixed before the question. We report $|P|$ in bytes and the actual number $B$ of visual tokens passed to the language model. The source cache and, when present, the model-input packet are reported separately. Packet sizes include headers and padding where present. API usage is reported in provider input tokens and includes prompt text.

Local experiments use frozen Qwen3-VL-4B-Instruct~\cite{qwen} in Q4\_K\_M with a Q8\_0 vision projector. The receiver reconstructs RGB frame pairs and uses the existing visual encoder, native spatial indices, and source-time coordinates. Every full $224\times224$ pair contributes 49 retained visual tokens. For compact readouts, we keep selected post-vision-encoder grid positions and count them at the language-model boundary. Local evaluation uses one question per inference. The main protocols use the visual stream and English multiple-choice questions with the original option order. Audio and subtitles are omitted from the main experiments.

\subsection{Event-aware readout at 392 tokens}

The sender processes the full video in one source traversal with the frame count known. It creates eight temporal bins and aggregates global RGB states using local translational alignment with an unaligned fallback. A second partition of the source timeline gives 14 event slots. Within each slot, the sender ranks adjacent-frame changes on a $7\times 7$ grid. The score adds the positive deviation from a running residual background to 0.2 times the current motion-compensated residual magnitude. Phase correlation estimates translation within four screen quadrants. The best-scoring pair retains its source RGB appearance and a local neighborhood around the most active cell.

The initial view occupies 25 tokens and the eight pooled views occupy 16 tokens each. Fourteen local event pairs share the remaining 239 tokens, with 17 tokens each except for the most salient event, which receives 18. The transport encodes RGB pairs in an H.264 payload with layout and time metadata. The receiver decodes these pairs and uses their selected grid positions. We vary encoder quality against the actual file length. The eight HERBench development clips use strict per-clip byte matching for the 392-token uniform and adaptive baselines. For the two new clips, the event packet is capped by the smallest same-budget baseline packet.

We alter this representation only at readout in the appearance and timing interventions. The appearance intervention replaces the bounding box of selected local cells with one RGB mean computed over the box and both frames. This preserves the patch's aggregate color, position, timestamps, global views, and token counts. The intervention removes spatial detail and within-pair variation in the box. The timing intervention rotates 13 same-sized event groups by six slots. It preserves event pixels and their original image coordinates, and keeps the unique 18-token event fixed. A further diagnostic rotates and resizes the local regions, jointly altering spatial appearance and temporal association. Each intervention uses the same underlying physical packet as the unmodified readout.

\subsection{Sharp-reference readout at 196 tokens}

This variant encodes a low-resolution predictive timeline and constructs four native frame pairs. At the start of each quarter, it compares a fresh intra reference, a correction relative to the transported prior reference, and reuse of the prediction where available. It ranks feasible choices by decoded squared RGB error within the current byte allowance. A signed local correction represents one transient excursion per quarter. Each readout pair binds the quarter reference to its corrected event state and retains the full 49-token grid. This construction is evaluated separately from the 392-token event layout.

\subsection{Regional refresh and dense temporal caches}

Regional refresh keeps a decoder-reproducible $224\times224$ image state. It scans every frame and tracks local discrepancies between source and reconstructed thumbnails on a $7\times7$ grid. The score smooths persistent error after discounting recent motion. At quarter-second opportunities, it writes a WebP-encoded 96-pixel square near the highest-scoring cell if error and byte allowance permit. Widespread discrepancy can trigger a whole-image refresh. Each record gives a source-frame index, region geometry, and encoded payload. Replay applies only records at or before the requested time. The whole-video readout uses 16 states organized as eight ordered pairs. Its 392-token budget is independent of the number of cache updates.

The dense variant pools every frame into its assigned time bin, at roughly up to four bins per second and about 1,100 bins per clip. It compares each pooled state to the last selected state via the maximum block discrepancy. States crossing the threshold are encoded with libx264. A timestamp index maps coded states to source bins, and replay holds the most recent available state across skipped bins. The encoder selects resolution, threshold, and quantization against a physical byte cap. A bin becomes available after the sender has seen all of its frames. The model sees 392 tokens in whole-video mode and can retrieve other times from the same packet in timestamped diagnostics.

\subsection{Evaluation protocols}

The event study uses ten full HERBench videos, with 28 official cross-segment questions per video. Eight videos were used during development. The frozen 392-token configuration was subsequently applied to two additional kitchen videos from participants absent from the development set. The longer pair lasts 15:32 and 32:58. A broader transfer cohort contains 84 videos and 610 questions, including these 280 HERBench items, 258 NExT-QA items, 36 MVBench items, and 36 Perception Test items~\cite{nextqa,mvbench,perception}. The broader study evaluates the 196-token representation and several selection baselines. MaxInfo Slow uses a port of the CLIP/MaxVol selection path. The coverage-only selector uses the corresponding FrameSelect implementation. The in-house adaptive baseline scans all frames and combines feature novelty, visual change, and temporal suppression before choosing whole frames.

The education and science cohort consists of 12 Video-MME public-knowledge clips (36 questions), 18 Video-MME-v2 diagram or board-work lessons (72 questions), and 15 SciVideoBench laboratory demonstrations (114 questions). Videos last 17.5~s to 47.3~min. In total, the encoder scans 640,594 source frames. We evaluate regional refresh against per-video byte-matched eight-frame WebP caches and 100-frame inputs. To match the eight-frame caches to regional refresh, the encoder adjusts quality and resolution and records fill bytes. All three matched-cache arms use 392 receiver tokens. The 100-frame baselines use 4,900 tokens for whole-video answering and follow a separate budget point.

For 75 questions across 15 clips that explicitly mention a time, all arms receive two or three views (98 or 147 tokens). A point uses the stated time and offsets of $\pm0.75$~s. A range uses three equally spaced interior observations. Five explicitly comparative questions use the two stated times. The 100-frame caches greedily return the nearest distinct retained views, including future neighbors if nearer. The source control uses frames at the queried times through the same model-input interface. The packet encoder never sees the questions.

\subsection{One-second change fidelity and statistics}

We sample 12 original one-second pairs per educational clip. A vision model screens the original pairs for recognizable new scientific content, without questions, answers, or method names. It marks 185 of 540 pairs as positive. Nine positives have zero measurable source difference under the common pixel protocol and are excluded. The screen identifies a descriptive subset. A separate attempt at semantic recovery grading fails its controls. The judge recognizes a copied original difference in only 14 of 31 trial videos and gives credit to held images in 12 of 31. We use the reproducible pixel metric for reconstruction comparisons.

For each positive pair, resize both source and reconstructed observations to $112\times112$ with area interpolation. In an $8\times8$ grid, keep the 16 cells with the largest source RGB difference energy. Let $D$ be the signed source RGB difference and $\widetilde D$ the reconstructed difference on those same cells. The fidelity score is
\[
  F(D,\widetilde D)=\frac{2\langle D,\widetilde D\rangle}
  {\|D\|_2^2+\|\widetilde D\|_2^2}.
\]
An exact, stationary, or reversed reconstruction gives scores of 1, 0, and $-1$, respectively, for every nonzero $D$. We report the mean score and the number of pairs with $F\geq 0.5$. The threshold is descriptive. For selected question-answer comparisons, we resample videos with replacement 10,000 times and recompute the paired difference in question-level accuracy. We report percentile 95\% intervals. Questions within a video are kept together. Comparisons and intervals are descriptive and unadjusted for multiplicity.

\section{Results}

\subsection{Cross-segment evidence and event appearance}

The event representation answers 98 of 280 HERBench questions, compared with 92 for uniform and 79 for adaptive selection at the same visual-token budget (Table~\ref{tab:her}). The paired difference is $+2.1$ percentage points against uniform (95\% interval $-1.4$ to $+6.1$) and $+6.8$ against adaptive (0.0 to $+12.9$). The two new kitchens contribute 23 of 56 correct for the event representation, 20 for uniform, and 17 for adaptive selection. Their mean event packet is 34,421 bytes. On these videos, the 196-token sharp-reference variant answers 20 questions at 14,037 mean bytes, matching the 20 correct of 392-token uniform selection at 34,614 bytes. At 196 tokens, uniform, adaptive, coverage-only, and MaxInfo answer 10, 19, 16, and 13 questions respectively. The sharp-reference packet uses 59.4\% fewer bytes than 392-token uniform selection on these two videos.

\begin{table}[htbp]
\centering
\caption{Correct answers on HERBench at 392 visual tokens. The combined column is descriptive. All intervention rows reuse the event packet for the corresponding video.}
\label{tab:her}
\begin{tabular}{@{}lrrr@{}}
\toprule
Input or intervention & Development (224) & New kitchens (56) & Total (280)\\
\midrule
Event representation & 75 & 23 & 98\\
Uniform frames & 72 & 20 & 92\\
Full-scan adaptive frames & 62 & 17 & 79\\
Local boxes reduced to mean RGB & 51 & 18 & 69\\
Local regions rotated and resized & 63 & 22 & 85\\
Event groups reassigned in time & 70 & 21 & 91\\
\bottomrule
\end{tabular}
\end{table}

\begin{figure}[htbp]
\centering
\includegraphics[width=\linewidth]{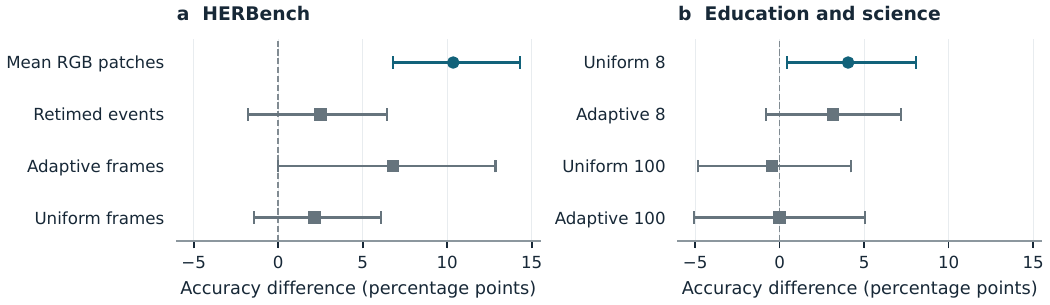}
\caption{Paired difference in question accuracy (percentage points) for the full representation minus the indicated comparison. Points show observed differences and bars show percentile 95\% intervals from video-cluster resampling. Panel (a) includes all ten HERBench videos, including eight development videos, with 280 questions. Panel (b) includes the 45 education and science videos, with 222 questions. The zero line marks equal observed accuracy.}
\label{fig:effects}
\end{figure}

Replacing local boxes by mean RGB yields 29 fewer correct answers. Keeping real local appearance changes 39 answers from wrong to right and ten from right to wrong at a fixed packet and receiver budget. The paired difference is $+10.4$ percentage points (95\% interval $+6.8$ to $+14.3$), with a positive raw difference on all ten videos. The two new kitchens account for six gains and one loss. Restoring the original event-to-time assignment yields seven additional correct answers ($+2.5$ points, 95\% interval $-1.8$ to $+6.4$). This contrast provides a less precise estimate of the role of the tested timing assignment (Fig.~\ref{fig:effects}a).

\subsection{Educational and scientific video question answering}

Regional refresh answers 66 of 222 questions (29.7\%) at 392 tokens (Table~\ref{tab:science}). The two byte-matched eight-frame baselines answer 57 and 59 correctly. The paired improvement is $+4.1$ percentage points against uniform (95\% interval $+0.4$ to $+8.1$) and $+3.2$ against adaptive ($-0.8$ to $+7.2$). The average physical cache length is 53,752 bytes. The subsequent regional-refresh model-input packet averages 24,152 bytes. The 100-frame baselines answer 67 and 66 correctly while using 12.5 times as many visual tokens. Their paired intervals both include zero (Fig.~\ref{fig:effects}b). The diagram subset favors regional refresh in raw counts. The laboratory subset favors the 100-frame baselines. These subgroup differences motivate testing which procedural details survive compression.

\begin{table}[htbp]
\centering
\caption{Correct answers on the 45 educational and scientific clips. Each eight-frame cache is matched to the regional cache's physical bytes per video. $B$ is the visual-token count per question in whole-video readout.}
\label{tab:science}
\small
\begin{tabular}{@{}lrrrrrr@{}}
\toprule
Subset & Questions & Regional & Uniform 8 & Adaptive 8 & Uniform 100 & Adaptive 100\\
\midrule
Visual tokens $B$ & & 392 & 392 & 392 & 4,900 & 4,900\\
All clips & 222 & 66 & 57 & 59 & 67 & 66\\
Public knowledge & 36 & 20 & 18 & 19 & 21 & 22\\
Diagram and board work & 72 & 25 & 22 & 21 & 20 & 19\\
Laboratory demonstrations & 114 & 21 & 17 & 19 & 26 & 25\\
\bottomrule
\end{tabular}
\end{table}

On the 75 timestamped questions, regional-cache replay answers 19 correctly, compared with 15 and 12 for nearest-time reads from the uniform and adaptive 100-frame caches. Direct source-frame lookup answers 18. Against adaptive retrieval, there are seven paired gains and zero losses ($+9.3$ percentage points, 95\% interval $+3.6$ to $+15.8$). Timestamped lookup is therefore useful under this particular readout and question selection. The counts leave considerable room for improving the evidence presented to the model.

\subsection{Reconstruction fidelity and answer accuracy}

\begin{figure}[htbp]
\centering
\includegraphics[width=\linewidth]{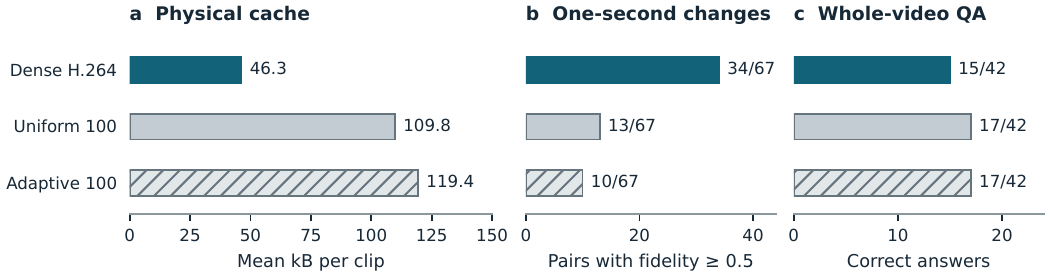}
\caption{Physical cache size, one-second change fidelity, and whole-video accuracy on the same 12 videos. Byte values are arithmetic means over clips in decimal kB. The pixel diagnostic contains 67 source-positive pairs with a measurable difference. Question answering contains 42 official items. The dense method uses 392 visual tokens, and the frame baselines use 4,900.}
\label{fig:dense}
\end{figure}

Across 176 one-second pairs, regional refresh reaches the fidelity threshold on 15 pairs, compared with zero and one for byte-matched uniform and adaptive eight-frame caches. Uniform and adaptive 100-frame caches reach it on 30 and 25 pairs. Their mean signed fidelities are 0.170 and 0.134, respectively, compared with 0.069 for regional refresh. On a fixed 12-clip subset (six science explainers and six diagram lessons), dense H.264 reaches the threshold on 34 of 67 pairs, compared with 13 and ten for the 100-frame caches. It averages 46,254 physical bytes per clip (Fig.~\ref{fig:dense}). Whole-video QA on those clips yields 15 of 42 correct for dense H.264 and 17 of 42 for each 100-frame method. In an additional same-packet diagnostic on these clips, a local before/after representation and the control that substitutes the before state for the after state both answer 13 questions correctly.

\subsection{Transfer, cloud readout, and diagnostic failures}

Over the broader 610-question transfer cohort, the 196-token sharp-reference variant answers 281 correctly. At 196 tokens, uniform, full-scan adaptive, coverage-only, and MaxInfo answer 291, 283, 302, and 297 respectively. The sharp-reference variant averages 10,374 bytes versus 10,558 for the four low-budget selectors. The coverage-only selector at 392 tokens answers 319 correctly. These results bound the interpretation of the two-kitchen operating point and show a setting where allocating a greater fraction of the budget to recognizable global frames helps.

On a 102-question subset of the education cohort, DeepSeek Flash answers 50 questions correctly with regional replay and 48 and 47 with uniform and adaptive 100-frame image inputs. Each cloud request batches the questions for one video. Regional replay averages 2,183 provider input tokens per video, compared with 19,544 for uniform 100-frame input. Mean successful request bodies are 92,679 and 779,008 bytes, respectively. An eight-frame uniform input is smaller still (2,060 tokens and 59,087 bytes per video) and answers 49 questions correctly. A text-only condition answers 38 questions correctly. These runs compare input formats within one service and protocol. They quantify input costs while providing limited evidence about the small observed accuracy differences.

On four additional HERBench movie trailers, we intervene on 12 questions with annotated evidence windows. In the 196-token sharp-reference arm, a source-level intervention replaces one evidence window with a preceding frame and re-encodes with the layout, byte count, and clear reference views controlled. The original, evidence-masked, and matched non-evidence-masked inputs answer nine, nine, and eight questions correctly. The explicit local readout touches only two of 12 masked evidence windows. In a separate repetition-counting test on 100 QUVA clips~\cite{quva}, the frozen model obtains zero exact counts from a decoded periodicity plot and one with a textual estimate. The standalone estimator obtains 19 exact counts from the same encoded trajectory. Both diagnostics expose a limit between what was processed or preserved and what the receiver can use.

\section{Discussion}

The same-packet appearance intervention offers a direct comparison of what the model receives from local events. The effect persists when physical bytes, timestamps, global context, and model tokens are fixed. Comparison against whole-frame uniform selection is weaker, and uncertainty about relative timing remains substantial. The degree to which these results extend beyond the tested camera viewpoints, event types, and model is open. In particular, eight HERBench videos participated in model-input design, and the two new long videos both show kitchen activity.

The educational cohort illustrates two uses of a time-indexed representation. It can provide compact whole-video input, with similar observed overall accuracy to 100-frame readout on this cohort. It can also replay a specific time without keeping a full-resolution frame bank. The H.264 result shows that stronger pixel-level retention can coexist with lower accuracy when the model reads only a few states. Better question-conditioned selection of windows and regions is a distinct problem from storing them at low cost.

Our cohorts are selected subsets of public datasets. They have heterogeneous option counts, question types, and evidence availability. Several subgroup and diagnostic comparisons were exploratory. The screen for original one-second changes is model-generated and was sampled sparsely. External benchmark leaderboards use different models, splits, input modalities, and budgets. The reported efficiencies cover cache transport and visual or provider tokens. End-to-end encoding latency, energy use, and peak memory were not isolated by method. The source must be decoded and inspected in all full-scan variants.

\section{Conclusion}

In this working study, retaining local event appearance improves a frozen model's answers in a same-packet HERBench intervention. Regional refresh provides useful time-indexed readout at a small model-input budget on educational videos. Dense time-indexed coding preserves short visual changes more efficiently on the 12-clip diagnosis. The latter gain does not carry through to whole-video accuracy under the tested readout. Evaluation of video caches for language models should report physical packet length, evidence fidelity, and answer accuracy together.

\section*{Data availability and tool disclosure}

The source benchmarks are identified in the references. Cohort definitions, per-question outputs, packet audits, and experiment scripts are retained in the study archive. The preprint source contains aggregate figures and includes no source video, benchmark question text, or API logs. OpenAI Codex assisted with experiment implementation, numerical cross-checks, and manuscript preparation. The vision and language models used for inference and diagnostic grading are described in Methods and Results.

\appendix
\section{Cohort construction and accounting details}

The 12 public-knowledge clips were selected by file size within topical and duration strata. The 18 diagram-oriented clips were selected by content among locally available Video-MME-v2 lessons before model scoring. The SciVideoBench subset takes the five most question-rich clips per selected discipline in the 180--900~s eligibility window, using media size and ID as tie-breakers. Its 15 clips cover physics, chemistry, and engineering labels. The 12-clip dense diagnostic uses the six explicit public science explainers and the first six diagram lessons in the saved cohort order. The 102-question cloud subset spans 21 videos. The study uses the official answer keys with exact option scoring. The sampled positive pairs and question subsets overlap with their parent cohorts. No pooled denominator combines them.

The byte-matched eight-frame cache comparison uses a mean of 48,493 effective bytes and 5,260 padding bytes for uniform selection, or 50,725 effective bytes and 3,027 padding bytes for adaptive selection, rounded separately. Median per-clip padding shares are 3.7\% and 3.4\%. The mean model-input packets for uniform and adaptive 100-frame inputs are 131,966 and 139,025 bytes on all 45 clips. The smaller subset used in Fig.~\ref{fig:dense} yields different means. In the broader transfer comparison, the sharp-reference packet stays at or below all four same-budget selection packets for every video. Byte gains are properties of these implementations and their specified operating points.

\end{document}